%% file: acl_latex.tex
\pdfoutput=1

\documentclass[11pt]{article}

\usepackage[final]{acl} 

\usepackage{times}
\usepackage{latexsym}

\usepackage[T1]{fontenc}

\usepackage[utf8]{inputenc}

\usepackage{microtype}

\usepackage{inconsolata}

\usepackage{graphicx}

\usepackage{graphicx} 

\usepackage{array} 
\usepackage{soul}

\usepackage{caption}

\usepackage{colortbl}
\usepackage{multirow}
\usepackage{dblfloatfix}

\title{GERestaurant: A German Dataset of Annotated Restaurant Reviews for Aspect-Based Sentiment Analysis}

\author{Nils Constantin Hellwig \\
  Media Informatics Group \\
  University of Regensburg \\
  Regensburg, Germany \\
  {\url{nils-constantin.hellwig@ur.de}} \And
  Jakob Fehle \\
Media Informatics Group \\
University of Regensburg \\
Regensburg, Germany \\
{\url{jakob.fehle@ur.de}} \\ \AND
  Markus Bink \\
Media Informatics Group \\
University of Regensburg \\
Regensburg, Germany \\
{\url{markus.bink@student.ur.de}} 
\And
Christian Wolff \\
Media Informatics Group \\
University of Regensburg \\
Regensburg, Germany \\
{\url{christian.wolff@ur.de}}}

\date{}

\begin{document}
\maketitle

\input{00_abstract.tex}
\input{01_introduction.tex}
\input{02_related_work}

\input{03_methodology}
\input{04_results.tex}
\input{05_limitations.tex}
\input{06_discussion.tex}
\input{07_conclusion.tex}
\input{08_ethical_considerations.tex}

\bibliography{bibliography}

\input{09_appendix.tex}

\end{document}

%% file: 00_abstract.tex
\begin{abstract}
We present GERestaurant, a novel dataset consisting of 3,078 German language restaurant reviews manually annotated for Aspect-Based Sentiment Analysis (ABSA). All reviews were collected from Tripadvisor, covering a diverse selection of restaurants, including regional and international cuisine with various culinary styles. The annotations encompass both implicit and explicit aspects, including all aspect terms, their corresponding aspect categories, and the sentiments expressed towards them. Furthermore, we provide baseline scores for the four ABSA tasks Aspect Category Detection, Aspect Category Sentiment Analysis, End-to-End ABSA and Target Aspect Sentiment Detection as a reference point for future advances. The dataset fills a gap in German language resources and facilitates exploration of ABSA in the restaurant domain. 
\end{abstract}

%% file: 01_introduction.tex
\section{Introduction}


Sentiment analysis (SA), also named opinion mining, is a research area in natural language processing (NLP) which involves the computational classification of individuals' sentiments, opinions and emotions. This usually involves categorizing sentiments into three polarities: positive, neutral and negative.


SA can be applied at both document- \cite{hellwig2023transformer,schmidt2022sentiment,tripathy2017document} and sentence-level \cite{liu2010sentiment}. However, if a document or sentence comprises a mixture of different sentiments, it's often impossible to assign a solely positive, negative or neutral label. As an example, consider the sentence "The salad tasted wonderful, but was quite expensive." of a restaurant review wherein positive sentiment is expressed towards the food while, concurrently, negative sentiment is expressed when addressing the food's price. To overcome this issue, Aspect-Based Sentiment Analysis (ABSA) has been extensively studied as it goes beyond assessing general sentiment and instead delves into a more granular examination of sentiment by linking particular aspects with corresponding sentiment polarities \cite{liu2005opinion,pontiki2015semeval}.

In this work, we introduce GERestaurant, a novel dataset comprising 3,078 German language restaurant reviews annotated for ABSA. It's the first German language dataset of sentences from restaurant reviews for ABSA. The annotations included the aspect term (if available), an aspect category selected from a predefined set of categories, and the sentiment or polarity expressed towards the aspect. The dataset is provided as a benchmark dataset for future research and parallels the widely used SemEval 2015 and 2016 restaurant datasets in terms of annotation scheme and annotation guidelines \cite{pontiki2015semeval, pontiki2016semeval}. Thus, it not only contributes to the availability of German language resources but also enables the exploration of new ABSA methods in the restaurant domain in the German language. Additionally, we provide a baseline performance by fine-tuning state-of-the-art (SOTA) transformer-based language models on the annotated dataset for typical ABSA tasks: Aspect Category Detection (ACD), Aspect Category Sentiment Analysis (ACSA), End-to-End ABSA (E2E-ABSA) and Target Aspect Sentiment Detection (TASD).


%% file: 02_related_work.tex
\section{Related Work}


ABSA has attracted increasing attention, in part due to benchmark datasets and shared tasks from various domains that facilitated the development of machine learning approaches for solving ABSA tasks. For instance, various datasets from different domains frequently employed in ABSA research include:

\begin{itemize}
\item \citet{ganu2009beyond}: A dataset comprising restaurant reviews in English, annotated with six pre-defined aspect categories assigned to sentiment polarities positive, neutral, negative, and conflict.
\item \citet{saeidi2016sentihood}: SentiHood, a dataset of English sentences extracted from a question answering (QA) platform discussing urban neighbourhoods. Annotations for aspect terms, their associated aspect categories, and the sentiment expressed towards them were provided.
\item \citet{jiang2019challenge}: MAMS, a dataset of English Tweets on celebrities, products, and companies. All aspect terms were annotated, along with the sentiment polarity expressed towards them.
\end{itemize}

However, the development of methods addressing the subtasks in ABSA was particularly driven by the SemEval shared task workshop in the years from 2014 to 2016 and the associated publishing of human-annotated datasets for ABSA. These comprised sentences from reviews of laptops and restaurants.

SemEval-2014 Task 4 \cite{pontiki2014semeval} was dedicated to ABSA and included annotations of aspect terms and the sentiment polarity expressed towards them. In addition, annotations of the aspect categories and the sentiment polarity expressed towards them are part of the provided dataset.

In the subsequent year, SemEval-2015 Task 12 \cite{pontiki2015semeval} was published, which included annotations of all aspect terms, their corresponding aspect category and the sentiment polarity expressed towards the aspect terms. Moreover, annotations of implicit aspects were provided, meaning cases where a sentiment was expressed towards an aspect category, without the presence of an aspect term. In such cases, the aspect term was annotated as \texttt{"NULL"}.

SemEval-2016 Task 5 \cite{pontiki2016semeval} encompassed the same three sentiment elements as SemEval-2015 Task 12 \cite{pontiki2015semeval}. In addition, subsets containing annotated sentences of hotel reviews and reviews in languages other than English were provided for each domain \cite{pontiki2015semeval}.

When examining datasets in German language, there is a scarcity of annotated datasets. The most prominent dataset in German language is the dataset published as part of the GermEval 2017 shared task \cite{wojatzki2017germeval}, which includes customer reviews concerning the "Deutsche Bahn", the German public train operator \cite{wojatzki2017germeval}. Reviews were annotated as a whole, rather than individual sentences separately \cite{wojatzki2017germeval}. Similar to the datasets introduced by \citet{pontiki2015semeval,pontiki2016semeval}, annotations were provided for all aspect terms, their associated aspect categories, and the sentiment expressed towards the aspect terms.

\citet{gabryszak2022mobasa} introduced the German language dataset MobASA, which comprises tweets from public transportation companies and channels related to barrier-free travel for handicapped passengers \cite{gabryszak2022mobasa}. Annotations covered aspect terms, their associated aspect categories, and the sentiments expressed towards each aspect term \cite{gabryszak2022mobasa}. 

In the realm of customer reviews, other notable resources include the SCARE corpus \cite{sanger2016scare}, comprising annotated application reviews from the Google Play Store, alongside annotations for aspect terms and sentiment polarities. Similarly, \citet{fehle2023aspect} introduced a dataset consisting of sentences from hotel reviews on Tripadvisor, whereby annotations are provided for the sentiments expressed towards the considered aspect categories.

%% file: 03_methodology.tex
\section{Methodology}
\begin{table*}[bp]
\scriptsize
\centering
\resizebox{2.0\columnwidth}{!}{
\begin{tabular}{lll}
\hline
\textbf{Aspect Category}                                             & \textbf{Triplets}                                                                                                                                                 & \textbf{Sentence}                                                                                                     \\ \hline
\begin{tabular}[c]{@{}l@{}}\texttt{GENERAL-}\\ \texttt{IMPRESSION}\end{tabular}        & \begin{tabular}[c]{@{}l@{}}{\texttt{[}}\texttt{('Restaurant', 'GENERAL-IMPRESSION',} \\ \texttt{'POSITIVE')}{\texttt{]}}\end{tabular}                                                                 & \textit{"Sehr schönes Restaurant."}                                                                                            \\ \arrayrulecolor[RGB]{200,200,200}\hline
\texttt{FOOD}                                                                 & {\texttt{[}}\texttt{('Bratwurst', 'FOOD', 'POSITIVE')}{\texttt{\texttt{]}}}                                                                                                                           & \textit{\begin{tabular}[c]{@{}l@{}}"Die Bratwurst war unglaublich lecker \\ und perfekt gewürzt."\end{tabular}}       \\ \hline
\texttt{SERVICE}                                                              & {\texttt{[}}\texttt{('Bedienung', 'SERVICE', 'NEGATIVE')}{\texttt{]}}                                                                                                                        & \textit{"Bedienung leider nicht aufmerksam."}                                                                                  \\ \hline
\texttt{AMBIENCE}                                                             & {\texttt{[}}\texttt{('NULL', 'AMBIENCE', 'NEGATIVE')}{\texttt{]}}                                                                                                                            & \textit{"Es war viel zu laut, wie im Club."}                                                                                   \\ \hline
\texttt{PRICE}                                                                & {\texttt{[}}\texttt{('NULL', 'PRICE', 'NEUTRAL')}{\texttt{]}}                                                                                                                                & \textit{"Preislich ist das ok gewesen."}                                                                              \\ \arrayrulecolor[RGB]{0,0,0}\hline
\begin{tabular}[c]{@{}l@{}}\texttt{PRICE},\\ \texttt{SERVICE}\end{tabular}                & \begin{tabular}[c]{@{}l@{}}{\texttt{[}}\texttt{('Preise', 'PRICE', 'NEUTRAL')}  \\  \texttt{('Service', 'SERVICE', 'NEGATIVE')}{\texttt{]}}\end{tabular}                                                   & \textit{"Preise sind ok und Service auch."}                                                                                    \\ \arrayrulecolor[RGB]{200,200,200}\hline
\begin{tabular}[c]{@{}l@{}}\texttt{FOOD},\\ \texttt{AMBIENCE}, \\ \texttt{SERVICE}\end{tabular} & \begin{tabular}[c]{@{}l@{}}{\texttt{[}}\texttt{('Essen', 'FOOD', 'POSITIVE')},  \\  \texttt{('Atmosphäre', 'AMBIENCE', 'POSITIVE'),  }\\  \texttt{('Service', 'SERVICE', 'POSITIVE')}{\texttt{]}}\end{tabular} & \begin{tabular}[c]{@{}l@{}}\textit{"Tolles Essen, tolle Atmosphäre und} \\ \textit{ganz netter und aufmerksamer Service!"}\end{tabular} \\ \arrayrulecolor[RGB]{0,0,0}\hline
\end{tabular}
}
\caption{Annotated examples for all aspect categories.}
\label{fig:annotation-examples}
\end{table*}
\subsection{Data Acquisition}

To gather German language restaurant reviews, Tripadvisor was selected as the data source. The five restaurants with the most customer reviews in the 25 most densely populated German cities as of 2022\footnote{German Federal Statistical Office, population and population density as of December 31, 2022: \url{https://www.destatis.de/DE/Themen/Laender-Regionen/Regionales/Gemeindeverzeichnis/Administrativ/05-staedte.html}} were considered, covering a wide spectrum of restaurant types, including regional and international cuisine with various culinary styles. In the course of the COVID-19 pandemic, restaurant reviews were influenced by the associated hygiene measures. To prevent sentiment bias introduced by hygiene regulations we included all reviews posted during a period without mandated COVID-19 hygiene restrictions, specifically reviews from October 15, 2022, to October 15, 2023, were taken into account.

Overall, a total of 3,212 user reviews with a German language label on Tripadvisor were collected. The reviews were segmented into 13,426 sentences using the NLTK Tokenizer \cite{loper2002nltk}. It was observed that, despite the German language code label, some sentences were in languages other than German. Due to this, \textit{langdetect}\footnote{\url{https://pypi.org/project/langdetect}} was employed to ascertain the language of each sentence, leading to the rejection of 631 sentences which resulted in a total of 12,795 sentences in German. 

Ultimately, the sentences underwent an anonymization process. Named entity recognition (NER) was employed using \textit{spaCy} (\textit{de\_core\_news\_lg} model) \cite{honnibal2017spacy} to replace locations, personal names, and time-related references with anonymized placeholders \textit{"LOC"}, \textit{"PERSON"} and \textit{"DATE"}. Subsequently, regular expressions were employed to substitute any mentions of the restaurant's name in the sentences with the placeholder \textit{"RESTAURANT\_NAME"}.

\subsection{Data Annotation}

From the complete set of 12,795 sentences, a subset of 5,000 sentences was randomly sampled for annotation. Care was taken to ensure an equal distribution of sentences from reviews with 1, 2, 3, 4 and 5-star ratings (1,000 sentences each) \footnote{A customer reviewing a restaurant on Tripadvisor is obligated to provide both a star rating and a textual assessment.}. This distribution was established so that each sentiment polarity occurs as evenly as possible across all sentences. 

\subsubsection{Annotation Task}

As proceeded for SemEval-2015 \cite{pontiki2015semeval} and SemEval-2016 \cite{pontiki2016semeval}, for a given sentence \textit{x}, one or multiple triplets \textit{(a, c, p)} should be assigned, where \textit{a} represents the aspect term, \textit{c} denotes the aspect category, and \textit{p} indicates the sentiment expressed towards the aspect. The annotations included the positional information of the aspect terms within the text. Multiple aspect terms could be assigned to the same aspect category. Similarly, an aspect term could be assigned to multiple aspect categories at once. Examples are presented in Table \ref{fig:annotation-examples} and an English language translation of the table is provided in Appendix \ref{appendix:annotation-examples-translation}. 

The four aspect categories \texttt{FOOD}, \texttt{SERVICE}, \texttt{AMBIENCE} and \texttt{PRICE} were considered, similar to the rating categories of the Zagat Survey \cite{lee2007incorporating} for restaurants. These categories can also be found on Tripadvisor, allowing users to optionally assign one to five stars to each category in addition to an overall star rating.

However, in contrast to the categories from the Zagat Survey and as preceded by \citet{pontiki2015semeval}, \texttt{AMBIENCE} was used as an aspect category instead of \textit{"Decor"} as it encompasses a slightly broader scope.  Furthermore, a fifth category called \texttt{GENERAL-IMPRESSION} was introduced, which captured aspects that pertain to the restaurant in a general sense, similar to the datasets for ABSA published in the realm of SemEval-2015 \cite{pontiki2015semeval} and SemEval-2016 \cite{pontiki2016semeval}, whereby an aspect category was introduced that encompassed general aspects related to a laptop or a restaurant for which a review was written. 

Implicit addressing of an aspect category should be annotated as well. In this case, \texttt{"NULL"} was assigned as the aspect term. For each aspect term within these categories, one of the following sentiment polarity labels should be applied: \texttt{POSITIVE}, \texttt{NEGATIVE}, \texttt{NEUTRAL} (indicating mild positivity or mild negativity sentiment) or \texttt{CONFLICT}. The \texttt{CONFLICT} label was assigned in case both positive and negative sentiments are expressed towards an aspect term.

Furthermore, as preceded by \citet{pontiki2015semeval}, aspects should only be annotated if a sentiment was expressed towards them. For instance, in the sentence \textit{"You can eat pizza there"}, no sentiment is expressed towards the aspect \textit{"Pizza"} (aspect category: \texttt{FOOD}), and thus, the aspect should not be annotated accordingly.

\subsubsection{Data Labelling Procedure}

Three persons were tasked with annotating sentences in order to establish the gold standard labels. Similar to the approach employed by \citet{pontiki2014semeval}, the annotation process commenced with one annotator (annotator \textit{A}, M.Sc. media computer science student) annotating all 5,000 sentences and subsequently, each of the annotations by annotator \textit{A} underwent inspection and validation by a second annotator \textit{B}. 

For the second annotation, a PhD student and an M.Sc. student, both specializing in media computer science, were tasked to review 2,500 annotations by annotator \textit{A} each. All annotators had prior experience in annotating textual data in the field of SA, with the PhD student having prior experience in annotating text for ABSA.

The annotation process was facilitated using \textit{LabelStudio}\footnote{Label Studio - Open Source Data Labelling Tool: \url{https://labelstud.io}}. All annotators were provided with a comprehensive annotation guideline\footnote{\url{https://github.com/NilsHellwig/GERestaurant/blob/main/annotation_guideline.pdf}
}, which explained the user interface in \textit{LabelStudio} specifically created for this annotation task and included examples for sentences in German language closely aligned with those provided in the annotation guideline employed by \citet{pontiki2015semeval}. 

In addition to annotating all triplets \textit{(a, c, p)}, annotators were tasked to tick a checkbox when they encountered two or more sentences in an example instead of one, since the NLTK Tokenizer employed for sentence segmentation could potentially introduce errors. Another checkbox was provided to mark examples where customers addressed an aspect without conveying any sentiment. This allowed for the possibility of annotating them at a later point in time for future studies.

In 113 out of 5,000 sentences, annotator \textit{B} proposed a label different to that assigned by annotator \textit{A}. Among these, Annotator \textit{A} accepted the revised label suggested by annotator \textit{B} in 81 sentences. The annotation of the remaining 32 sentences was decided in consensus of the two annotators. In 16 sentences, both annotators opted to adopt the annotations provided by annotator \textit{A}, in seven instances, the annotation of annotator \textit{B} was adhered to. For the remaining nine sentences, a consensus was reached on an annotation distinct from their initially proposed annotations.

Among the 5,000 examples, 589 were excluded since they consisted of more than one sentence. Subsequently, out of the remaining 4,411 sentences, 1,291 were omitted since no sentiment was expressed towards aspects of the considered aspect categories and 42 sentences were removed since they encompassed at least one triplet with a conflict polarity, resulting in a total of 3,078 sentences with a total of 3,149 explicit and 1,165 implicit aspects.

\subsection{Baseline Models}

For a total of four typical ABSA tasks, we provide transformer-based baseline models. All models were loaded using the Hugging Face \textit{transformers} library\footnote{\url{https://pypi.org/project/transformers}} and trained on two NVIDIA RTX A5000 GPU with 24 GB VRAM each. The implementation was conducted using Python version 3.11.5. To assess the performance of each model, we conducted a random 70-30 train-test split. The models were trained on the training set, consisting of 2,154 examples, and evaluated on the test set, containing 924 examples. 

\subsubsection{Aspect Category Detection (ACD) and Aspect Category Sentiment Analysis (ACSA)}

Similar to \citet{fehle2023aspect}, the identification of aspect categories (ACD) and the identification of both aspect categories and the sentiment polarity expressed towards them (ACSA) was treated as a multi-label classification task. Two base models were fine-tuned in this study: gbert-large\footnote{\url{https://huggingface.co/deepset/gbert-large}} (337 million parameters) and gbert-base\footnote{\url{https://huggingface.co/deepset/gbert-base}} (111 million parameters) by \textit{deepset}. Both models are based on the BERT architecture and are pre-trained on large amounts of German language texts \citep{chan2020german}.

For training and validation, a batch size of 16, an epoch-number of 3 and a learning rate of 2e-5 (c.f. \citet{devlin2018bert}) was used. As proceeded by \citet{fehle2023aspect}, a prediction was considered a true positive, if the predicted aspect(s) of a sentence (including the sentiment polarity for ACSA) occurred in the ground truth labels.

\subsubsection{End-to-End ABSA (E2E-ABSA)}

E2E-ABSA is the task that aims at simultaneously identifying aspect terms and determining the sentiment polarity expressed towards them in a given text. As proceeded by \citet{li2019exploiting}, E2E-ABSA was conducted employing a BERT model for token classification. gbert-large and gbert-base were employed for this task as well.  The task involved predicting a tag sequence \( y = \{y_1, \ldots, y_T\} \), with each tag corresponding to a token in the sentence. The potential values for \( y_t \) encompass B-\{\(POS\), \(NEG\), \(NEU\)\}, I-\{\(POS\), \(NEG\), \(NEU\)\} or O. The tag denoted the beginning (B) and inside (I) of an aspect term, coupled with negative, neutral or positive sentiment and O, in case that a token was not a part of an aspect term.

For training, a binary cross-entropy loss was employed, and the sigmoid function was used as the activation function. Similar to the  evaluations conducted by \citet{li2019exploiting}, learning rate was set to 2e-5, batch size was set to 16 and the model was trained for 1,500 steps. When calculating the evaluation metrics, the true positives included all correctly identified pairs of an aspect term and the sentiment polarity expressed towards it, similar to \citet{zhang2023sentiment} and \citet{li2019exploiting}.

\subsubsection{Target Aspect Sentiment Detection (TASD)}

TASD is the task that leverages the full complexity of GERestaurants' annotations. Its objective is to identify all aspect terms, their associated aspect categories, and the sentiment expressed towards the aspect terms within a given text. 

For the TASD task, the paraphrasing approach methodology introduced by \citet{zhang2021aspect} was employed. The paraphrase generation framework utilized is outlined in Appendix \ref{appendix:paraphrase-generation-framework}. The polarity label \texttt{POSITIVE} was mapped to \textit{“gut"} (Eng.: \textit{"good"}) in the paraphrased label, \texttt{NEGATIVE} to \textit{"schlecht"} (Eng.: \textit{"bad"}) and \texttt{NEUTRAL} to \textit{"ok"}. In the case of an implicit aspect, the aspect term was decoded as \textit{"es"} (Eng.: \textit{"it"}).

Both t5-large\footnote{\url{https://huggingface.co/t5-large}} (770 million parameters) and t5-base\footnote{\url{https://huggingface.co/t5-base}} (223 million parameters) were evaluated as the underlying seq2seq models. In terms of training parameters, batch size was set to 16, number of training epochs to 20 and learning rate to 3e-4, similar to \citet{zhang2021aspect}. For evaluation, true positives encompassed all correctly identified triplets, meaning that all three sentiment elements (aspect term, aspect category and sentiment polarity) were identified correctly.

%% file: 04_results.tex
\section{Results}

\subsection{Properties of the Annotated Dataset}

\begin{table*}[!h]
\scriptsize
\centering
\resizebox{2.0\columnwidth}{!}{
\begin{tabular}{lrrrrrrrr}
\hline
\multicolumn{1}{c}{\textbf{}}                & \multicolumn{2}{c}{\textbf{Positive}}                                         & \multicolumn{2}{c}{\textbf{Negative}}                                         & \multicolumn{2}{c}{\textbf{Neutral}}                                          & \multicolumn{2}{c}{\textbf{Total}}                                            \\ \hline
\multicolumn{1}{c}{\textbf{Aspect Category}} & \multicolumn{1}{c}{\textbf{Explicit}} & \multicolumn{1}{c}{\textbf{Implicit}} & \multicolumn{1}{c}{\textbf{Explicit}} & \multicolumn{1}{c}{\textbf{Implicit}} & \multicolumn{1}{c}{\textbf{Explicit}} & \multicolumn{1}{c}{\textbf{Implicit}} & \multicolumn{1}{c}{\textbf{Explicit}} & \multicolumn{1}{c}{\textbf{Implicit}} \\ \hline
\begin{tabular}[c]{@{}l@{}}\texttt{GENERAL-}\\ \texttt{IMPRESSION}\end{tabular} & 103                                   & 306                                   & 56                                    & 285                                   & 5                                     & 21                                    & 164                                   & 612                                   \\
\texttt{FOOD}                                                          & 880                                   & 83                                    & 532                                   & 98                                    & 109                                   & 10                                    & 1,521                                 & 191                                   \\
\texttt{SERVICE}                                                       & 514                                   & 69                                    & 316                                   & 177                                   & 10                                    & 0                                     & 840                                   & 246                                   \\
\texttt{AMBIENCE}                                                      & 312                                   & 26                                    & 99                                    & 42                                    & 6                                     & 0                                     & 417                                   & 68                                    \\
\texttt{PRICE}                                                         & 45                                    & 1                                     & 149                                   & 41                                    & 13                                    & 6                                     & 207                                   & 48                                    \\ \hline
\textbf{Total}                                                & 1,854                                 & 485                                   & 1,152                                 & 643                                   & 143                                   & 37                                    & 3,149                                 & 1,165                                 \\ \hline
\end{tabular}
}
\caption{Aspect categories distribution per sentiment polarity and reference type for the annotated dataset.}
\label{fig:total-dataset-statistics}
\end{table*}

Table \ref{fig:total-dataset-statistics} presents an overview of the frequency of triplets occurring with their respective aspect categories, reference types, and sentiment polarities in the overall dataset. The highest number of triplets was identified for the \texttt{FOOD} category (1,712 triplets), while the lowest count was observed for the \texttt{PRICE} category (255 triplets). Aspects were more frequently addressed explicitly (3,149 triplets) rather than implicitly (1,165 triplets). Positive sentiments were expressed towards the majority of identified aspects (2,339 triplets), followed by negative sentiments (1,795 triplets). A neutral sentiment was expressed towards 180 aspects.

\begin{table*}[h]
\scriptsize
\centering
\resizebox{2.0\columnwidth}{!}{
\begin{tabular}{lll}
\hline
\textbf{Aspect Category}    & \textbf{Description}                                                                                                                                                                                                    & \textbf{Most Frequent Aspect Terms}                                                                                           \\ \hline
\texttt{GENERAL-IMPRESSION} & \begin{tabular}[c]{@{}l@{}}Aspects related to the overall impression of the \\ restaurant without focusing on the aforementioned \\ aspect categories.\end{tabular}                                                     & \begin{tabular}[c]{@{}l@{}}\textit{Restaurant} (42)\\ \textit{RESTAURANT\_NAME} (22)\\ \textit{LOC} (22)\\ \textit{Lokal} (12)\\ \textit{Brauhaus} (5)\end{tabular}        \\ \arrayrulecolor[RGB]{200,200,200}\hline
\texttt{FOOD}               & \begin{tabular}[c]{@{}l@{}}Aspects related to food in general or specific \\ dishes and drinks.\end{tabular}                                                                                                            & \begin{tabular}[c]{@{}l@{}}\textit{Essen} (302)\\ \textit{Bier} (46)\\ \textit{Speisen} (42)\\ \textit{Fleisch} (30)\\ \textit{Küche} (28)\end{tabular}                    \\ \arrayrulecolor[RGB]{200,200,200}\hline
\texttt{SERVICE}            & \begin{tabular}[c]{@{}l@{}}Aspects related to service in general or the \\ attitude and professionalism of staff, wait times, \\ or service offerings such as takeout.\end{tabular}                                     & \begin{tabular}[c]{@{}l@{}}\textit{Service} (209)\\ \textit{Bedienung} (125)\\ \textit{Personal} (90)\\ \textit{Kellner} (58)\\ \textit{Kellnerin} (17)\end{tabular}       \\ \arrayrulecolor[RGB]{200,200,200}\hline
\texttt{AMBIENCE}           & \begin{tabular}[c]{@{}l@{}}Aspects related to the ambiance and atmosphere \\ in general or the environment of the restaurant's \\ interior and exterior, including its decor and \\ entertainment options.\end{tabular} & \begin{tabular}[c]{@{}l@{}}\textit{Ambiente} (103)\\ \textit{Atmosphäre} (51)\\ \textit{Lage} (13)\\ \textit{Lokal} (12)\\ \textit{Location} (10)\end{tabular}             \\ \arrayrulecolor[RGB]{200,200,200}\hline
\texttt{PRICE}              & \begin{tabular}[c]{@{}l@{}}Aspects related to price in general or the pricing \\ of dishes, beverages, or other services offered \\ by the restaurant.\end{tabular}                                                     & \begin{tabular}[c]{@{}l@{}}\textit{Preise} (30)\\ \textit{Preis} (25)\\ \textit{Essen} (14)\\ \textit{Preisen} (11)\\ \textit{Preis-Leistungsverhältnis} (10)\end{tabular} \\ \arrayrulecolor[RGB]{0,0,0}\hline
\end{tabular}
}
\caption{Description of the aspect categories and their most frequent aspect terms.}
\label{fig:most-frequent-aspect-terms}
\end{table*}

\begin{table}[h]
\scriptsize
\centering
\resizebox{1.0\columnwidth}{!}{
\begin{tabular}{ccccccccccc}
\hline
\textbf{\# Triplets}   & 1     & 2   & 3   & 4  & 5  & 6 & 7 & 8 & 9 & 16 \\
\textbf{\# Sentences}  & 2,236 & 590 & 168 & 57 & 14 & 7 & 3 & 1 & 1 & 1  \\ \hline
\end{tabular}
}
\caption{Sample count of each triplet quantity.}
\label{tab:triplet-frequency-statistics}
\end{table}

Moreover, Table \ref{fig:most-frequent-aspect-terms} presents the most frequently occurring aspect terms within each aspect category, and Table \ref{tab:triplet-frequency-statistics} shows the sample count for each triplet quantity. In the case of all aspect categories except for \texttt{GENERAL-IMPRESSION}, the most frequently occurring aspect term is equal to the name of the corresponding aspect category. Moreover, in more than two-thirds (2,236) of the 3,078 sentences, exactly one aspect or triplet was identified.

\setlength\tabcolsep{3pt}

\begin{table*}[h]
\scriptsize
\centering
\resizebox{2.0\columnwidth}{!}{
\begin{tabular}{lcccccccccc}
\hline
\multicolumn{1}{c}{\multirow{2}{*}{\textbf{Dataset}}} &
  \multicolumn{5}{c}{\multirow{2}{*}{\textbf{Aspect Category}}} &
  \multicolumn{3}{c}{\multirow{2}{*}{\textbf{Polarity}}} &
  \multicolumn{2}{l}{\multirow{2}{*}{\textbf{Aspect Term Type}}} \\
\multicolumn{1}{c}{} & \multicolumn{5}{c}{}                                           & \multicolumn{3}{c}{}                         & \multicolumn{2}{l}{} \\ \hline
 &
  \textbf{\begin{tabular}[c]{@{}c@{}}General \\ Impression\end{tabular}} &
  \textbf{Food} &
  \textbf{Service} &
  \textbf{Ambience} &
  \multicolumn{1}{c|}{\textbf{Price}} &
  \textbf{Positive} &
  \textbf{Negative} &
  \multicolumn{1}{c|}{\textbf{Neutral}} &
  \textbf{Implicit} &
  \textbf{Explicit} \\ \hline
GERestaurant         & 18.0\% & 39.7\% & 25.2\% & 11.2\% & \multicolumn{1}{c|}{5.9\%} & 54.2\% & 41.6\% & \multicolumn{1}{c|}{4.2\%} & 27.0\%    & 73.0\%   \\
SemEval 2015         & 20.6\% & 42.6\% & 17.7\% & 11.5\% & \multicolumn{1}{c|}{7.5\%} & 66.1\% & 30.0\% & \multicolumn{1}{c|}{3.9\%} & 24.9\%    & 75.1\%   \\
SemEval 2016         & 20.6\% & 43.6\% & 17.9\% & 10.8\% & \multicolumn{1}{c|}{7.1\%} & 67.4\% & 28.3\% & \multicolumn{1}{c|}{4.3\%} & 24.8\%    & 75.2\%   \\

\hline
\end{tabular}
}

\caption{Comparison of the balances of the aspect category, the polarity labels and the ratio of implicitly and explicitly expressed aspect terms between the three ABSA datasets GERestaurant, SemEval 2015 and SemEval 2016.}
\label{fig:comparison-class-balance}
\end{table*}

\setlength\tabcolsep{5pt}

\subsection{Comparison with the SemEval Datasets}

As the dataset used in this work and the datasets from SemEval 2015 and 2016 are similar in terms of their domain and the type and depth of annotation, it is possible to compare dataset properties, such as their class distribution or language-specific features, such as the ratio of explicitly and implicitly expressed aspects.
In order to ensure the comparability of the annotations of the GERestaurant dataset with the two SemEval datasets from 2015 and 2016, various adjustments had to be made, as although the datasets have undergone similar annotation procedures, the labels of the aspect categories are named and summarized differently: (1) The \texttt{PRICES} subcategories of the SemEval datasets were transformed to the \texttt{PRICE} aspect category; (2) the \texttt{RESTAURANT} category of the SemEval datasets was converted to the \texttt{GENERAL-IMPRESSION} category; (3) the \texttt{LOCATION} category of the SemEval datasets were integrated into the \texttt{AMBIENCE} category; and (4) The \texttt{DRINKS} category of the SemEval datasets was merged into the \texttt{FOOD} category.

Table \ref{fig:comparison-class-balance} depicts the class balances of the five aspect categories as well as the polarity labels over the three datasets GERestaurant, SemEval 2015 and 2016. Subsequently, we consider a dataset as the combination of its train and test sets. The balance of the aspect classes of the SemEval datasets is almost identical, facilitated in part by the fact that almost the entire SemEval 2015 dataset, with 1,700 of its 1,702 annotated examples, has been integrated into the SemEval 2016 dataset, which contains a total of 2,384 annotated examples. The overall class distribution of the GERestaurant dataset is also quite similar to that of the SemEval datasets and differs primarily in a 6.5 percentage point higher occurrence of the \texttt{SERVICE} aspect category, while all other aspect classes occur slightly less frequently.
Considering the distributions of the polarity classes across all aspects, while the overall distributions of the polarity labels between the SemEval datasets are again very similar, bigger differences can be observed between the GERestaurant and SemEval datasets. The proportion of the neutral label  remains comparably low between all datasets, but the negative polarity label was assigned up to 12 percentage points more frequently in the GERestaurant dataset at 41.6\%, while the positive label was correspondingly annotated less frequently compared to the SemEval datasets, constituting only 54.5\% of the total.
Similar to the distribution of aspect classes, the ratio of implicitly and explicitly expressed aspects is very similar between all corpora. While the two SemEval datasets have an almost identical ratio, the GERestaurant dataset is only slightly above in terms of implicit aspects with an increase of about two percentage points, resulting in 27.0\% implicitly expressed aspects and 73.0\% explicitly expressed aspects.

\subsection{Baseline Performance}

The performance achieved in the four ABSA tasks under consideration are presented in Table \ref{fig:baseline-performance}. For predicting the five aspect classes (ACD task), gbert-large demonstrated the highest performance, achieving micro and macro F1 scores of 91.82 and 90.73, respectively, placing it approximately three percentage points ahead of gbert-base. Similarly, in the classification of aspects combined with their polarity (ACSA), the best performance was observed when employing gbert-large, which attained micro and macro F1 scores of 85.14 and 58.61, respectively. The micro-averaged F1 score surpassed that achieved with gbert-base by approximately 11 percentage points, while in the case of the macro-averaged F1 score, it exceeded it by around 22 percentage points.

\begin{table}[h]
\scriptsize
\centering
\resizebox{1.0\columnwidth}{!}{
\begin{tabular}{llcc}
\hline
\textbf{Task}                      & \textbf{Language Model} & \multicolumn{1}{c}{\textbf{F1 Micro}} & \multicolumn{1}{c}{\textbf{F1 Macro}} \\ \hline
\multirow{2}{*}{ACD} & gbert-large & 91.82 & 90.73 \\
  & gbert-base & 88.76 & 87.82 \\
\arrayrulecolor{gray}\hline\arrayrulecolor{black}
\multirow{2}{*}{ACSA} & gbert-large & 85.14 & 58.61 \\
  & gbert-base & 73.85 & 36.63 \\
\arrayrulecolor{gray}\hline\arrayrulecolor{black}
\multirow{2}{*}{E2E-ABSA} & gbert-large & 81.61 & 77.28 \\
  & gbert-base & 74.66 & 50.25 \\
  \arrayrulecolor{gray}\hline\arrayrulecolor{black}
\multirow{2}{*}{TASD} & t5-large & 68.86 & 59.03 \\
  & t5-base & 64.74 & 54.32 \\
\hline
\end{tabular}
}

\caption{Performance for the baseline models per ABSA subtask.}
\label{fig:baseline-performance}
\end{table}

For the E2E-ABSA task, gbert-large demonstrated the highest performance as well, achieving a micro F1 score of 81.61 and a macro F1 score of 77.28. This performance improvement over gbert-base, with a micro F1 score of 74.66 and a macro F1 score of 50.25.

Similarly to the previous tasks, again, the large model variant exceeded the performance of the base model by about four to five percentage points, resulting in a micro F1 score of 68.86 a macro F1 score of 59.03 for the t5-large model.

%% file: 05_limitations.tex
\section{Limitations}

While GERestaurant provides a valuable resource for studying ABSA in the German restaurant domain, it also comes with limitations. Firstly, the annotations are based on human judgments, which introduces subjectivity and potential inconsistencies. Furthermore, the quality of annotations is constrained by the fact that each example was not independently annotated by multiple annotators, but rather, one annotator annotated all sentences and their annotations were reviewed by another annotator.

Furthermore, the imbalance among the five aspect categories can be considered a limitation of this work. For instance, the fewest number of aspects (251) are assigned to the \texttt{PRICE} category, while the majority of aspects (1,676) are assigned to the \texttt{FOOD} category. Similar imbalances are observed in terms of sentiment polarities, with only 175 aspects toward which a neutral sentiment was expressed, compared to 2,283 aspects towards which a positive sentiment was expressed, which represents more than half of all aspects.

%% file: 06_discussion.tex
\section{Discussion}
GERestaurant offers a novel resource for ABSA research in the German language, specifically within the restaurant domain. Comprising 3,078 manually annotated sentences, GERestaurant encompasses both implicit and explicit aspects, annotated by human annotators. This is the third German language dataset besides GermEval 2017 \citep{wojatzki2017germeval} and MobASA \citep{gabryszak2022mobasa} to include annotations of aspect terms, aspect categories, and sentiment polarities of both implicit and explicit aspects. 

The analysis of the class distributions of the aspect classes and the sentiment polarities between the German GERestaurant dataset and the English SemEval 2015 and 2016 datasets revealed a strong similarity of the ABSA-specific annotations of the datasets. The close correlation between the datasets opens up a variety of possibilities to compare the performance of ABSA methods on English and German datasets and could provide conclusions on how far methods can be used across languages despite language-specific differences in the datasets and methods.

Our provided baseline performance on all four ABSA tasks is in line with the performance reported in similar studies using transformer-based models for such tasks across various domains. However, it's important to acknowledge that the comparability of the results is limited due to variations in the number of aspect categories and the number of training examples across the datasets.  

A micro-averaged F1 score of 91.82 was achieved in the ACD task, consistent with micro-averaged F1 scores obtained on other datasets, e.g. a micro-averaged F1 score of 90.89 on the restaurant dataset of SemEval from 2014 \citep{sun2019utilizing} or a micro-averaged F1 score of 90.6 on the dataset comprising hotel reviews presented by \citet{fehle2023aspect}. 

In the ACSA task, a micro-averaged F1 score of 85.14 was obtained, slightly exceeding the reported scores achieved on other datasets. \citet{cai2020aspect} reported micro-averaged F1 scores of 64.67 and 74.55 for the restaurant datasets of SemEval 2015 and 2016, respectively. \citet{assenmacher2021re} reported a micro-averaged F1 score of 65.5 on GermEval 2017 and \citet{fehle2023aspect} reported a micro-averaged F1 score of 80.9 on the dataset comprising hotel reviews. 

For the E2E-ABSA task, a micro-averaged F1 score of 81.61 was attained. Lower scores were reported for other domains, e.g. \citet{li2019exploiting} reported a micro-averaged F1 score of 73.22 when considering the restaurant domain and 60.43 when considering the laptop domain, using datasets composed of examples from the SemEval datasets from 2014 to 2016. 

The performance in the TASD task (micro-averaged F1 score of 68.86) falls within the spectrum of results observed by \citet{zhang2021aspect}, who represented triplets as phrases, reporting a micro-averaged F1 score of 63.06 for the restaurant dataset of SemEval 2015 and a micro-averaged F1 score of 71.97 for the restaurant dataset of SemEval 2016.

%% file: 07_conclusion.tex
\section{Conclusion \& Future Work}

This work presents GERestaurant, a novel German language dataset comprising 3,078 restaurant reviews annotated for ABSA. The dataset covers implicit and explicit aspects, providing annotations for aspect terms, aspect categories, and sentiment polarities. Transformer-based SOTA models were fine-tuned on the training set provided by us for four common ABSA tasks, and subsequently evaluated on the test set.

In future work, GERestaurant could be utilized for developing improved machine learning models with focus on the German language for various ABSA tasks, building upon the methods introduced in this work and further improving the presented baseline values. Moreover, future work may involve expanding the aspect categories by incorporating fine-grained attributes, as in the SemEval datasets from 2015 and 2016, or including information about not only aspect phrases but also opinion phrases, in order to reflect the entire quadruple of an aspect-based annotation \cite{pontiki2015semeval,pontiki2016semeval}. 

%% file: 08_ethical_considerations.tex
\section{Ethical Considerations}

The collection of our dataset adhered to strict privacy guidelines to safeguard the rights of users. Our primary objective was to extract reviews while avoiding the collection of personalized data that could potentially identify individual users or specific user groups. Furthermore, any direct references to individuals or restaurants were systematically anonymized to prevent indirect identification of individuals or establishments. 

The dataset and its annotations are available upon request from the authors to ensure responsible usage for academic purposes, thus preserving the original intent of data collection. The Python code for data collection and data cleaning is accessible via GitHub\footnote{\url{https://github.com/NilsHellwig/GERestaurant}}.

Despite our meticulous data collection and anonymization procedures, inherent limitations and ethical considerations persist. Our dataset may not fully represent the spectrum of user sentiment due to potential bias in review writing, as reviewers may only represent a specific subset of the population. Furthermore, the transferability of knowledge about review semantics and characteristics across different rating platforms cannot be guaranteed.

%% file: 09_appendix.tex
\onecolumn
\appendix

\section{Appendix}

\subsection{Examples from the Annotated Dataset}\label{appendix:annotation-examples-translation}

\begin{table}[h]
\scriptsize
\centering
\resizebox{1.0\columnwidth}{!}{
\begin{tabular}{lll}
\hline
\textbf{Aspect Category}                                             & \textbf{Triplets}                                                                                                                                                 & \textbf{Sentence}                                                                                                     \\ \hline
\begin{tabular}[c]{@{}l@{}}\texttt{GENERAL-}\\ \texttt{IMPRESSION}\end{tabular}        & \begin{tabular}[c]{@{}l@{}}{\texttt{[}}\texttt{('restaurant', 'GENERAL-IMPRESSION',} \\ \texttt{'POSITIVE')}{\texttt{]}}\end{tabular}                                                                 & \textit{"Very nice restaurant."}                                                                                            \\ \arrayrulecolor[RGB]{200,200,200}\hline
\texttt{FOOD}                                                                 & {\texttt{[}}\texttt{('sausage', 'FOOD', 'POSITIVE')}{\texttt{]}}                                                                                                                           & \textit{\begin{tabular}[c]{@{}l@{}}"The sausage was incredibly delicious \\ and perfectly seasoned."\end{tabular}}       \\ \hline
\texttt{SERVICE}                                                              & {\texttt{[}}\texttt{('Service', 'SERVICE', 'NEGATIVE')}{\texttt{]}}                                                                                                                        & \textit{"Service unfortunately not attentive."}                                                                                  \\ \hline
\texttt{AMBIENCE}                                                             & {\texttt{[}}\texttt{('NULL', 'AMBIENCE', 'NEGATIVE')}{\texttt{]}}                                                                                                                            & \textit{"It was much too loud, like in a club."}                                                                                   \\ \hline
\texttt{PRICE}                                                                & {\texttt{[}}\texttt{('NULL', 'PRICE', 'NEUTRAL')}{\texttt{]}}                                                                                                                                & \textit{"Price-wise it was ok."}                                                                              \\ \arrayrulecolor[RGB]{0,0,0}\hline
\begin{tabular}[c]{@{}l@{}}\texttt{PRICE},\\ \texttt{SERVICE}\end{tabular}                & \begin{tabular}[c]{@{}l@{}}{\texttt{[}}\texttt{('Prices', 'PRICE', 'NEUTRAL')}  \\  \texttt{('service', 'SERVICE', 'NEGATIVE')}{\texttt{]}}\end{tabular}                                                   & \textit{"Prices are ok and service as well."}                                                                                    \\ \arrayrulecolor[RGB]{200,200,200}\hline
\begin{tabular}[c]{@{}l@{}}\texttt{FOOD},\\ \texttt{AMBIENCE}, \\ \texttt{SERVICE}\end{tabular} & \begin{tabular}[c]{@{}l@{}}{\texttt{[}}\texttt{('food', 'FOOD', 'POSITIVE')},  \\  \texttt{('atmosphere', 'AMBIENCE', 'POSITIVE'),  }\\  \texttt{('service', 'SERVICE', 'POSITIVE')}{\texttt{]}}\end{tabular} & \begin{tabular}[c]{@{}l@{}}\textit{"Great food, great atmosphere and} \\ \textit{really nice and attentive service!"}\end{tabular} \\ \arrayrulecolor[RGB]{0,0,0}\hline
\end{tabular}
}
\caption{Annotated examples for all aspect categories (English translation).}
\label{fig:annotation-examples-translation}
\end{table}

\subsection{Paraphrase Generation Framework}\label{appendix:paraphrase-generation-framework}

\definecolor{darkgreen}{RGB}{0,128,0}
\definecolor{darkred}{RGB}{230,0,0}

\subsubsection{Explicit Aspect}

\begin{table}[h]
\scriptsize
\centering
\begin{tabular}{p{3cm}p{7cm}}
\hline
\textbf{Sentence (Input)}  & \texttt{Die \textcolor{blue}{Pasta} war super, aber die \textcolor{blue}{Bedienung} war unfreundlich!}                                                \\ \cline{2-2} 
\textbf{Label}             & \begin{tabular}[c]{@{}l@{}}\texttt{{[}(\textcolor{blue}{'Pasta'}, \textcolor{orange}{'FOOD'}, \textcolor{darkgreen}{'POSITIVE'}),}\\ \texttt{(\textcolor{blue}{'Bedienung'}, \textcolor{orange}{'SERVICE'}, \textcolor{red}{'NEGATIVE'}){]}}\end{tabular}   \\ \cline{2-2} 
\textbf{Paraphrased Label} & \begin{tabular}[c]{@{}l@{}}\texttt{\textcolor{orange}{Essen} ist \textcolor{darkgreen}{gut}, weil \textcolor{blue}{Pasta} \textcolor{darkgreen}{gut} ist {[}SSEP{]}} \\ \texttt{\textcolor{orange}{Service} ist \textcolor{darkred}{schlecht}, weil \textcolor{blue}{Bedienung} \textcolor{darkred}{schlecht} ist {[}SSEP{]}}\end{tabular} \\ \hline
\end{tabular}
\caption{Paraphrasing of an explicit aspect's label.}
\end{table}

\subsubsection{Implicit Aspect}

\begin{table}[h]
\scriptsize
\centering
\begin{tabular}{p{3cm}p{7cm}}
\hline
\textbf{Sentence (Input)}  & \texttt{Es hat richtig gut geschmeckt!} \\ \cline{2-2} 
\textbf{Label}             & \texttt{{[}(\textcolor{blue}{'NULL'}, \textcolor{orange}{'FOOD'}, \textcolor{darkgreen}{'POSITIVE'}){]}}   \\ \cline{2-2} 
\textbf{Paraphrased Label} & \texttt{\textcolor{orange}{Essen} ist \textcolor{darkgreen}{gut}, weil \textcolor{blue}{es} \textcolor{darkgreen}{gut} ist {[}SSEP{]}} \\ \hline
\end{tabular}
\caption{Paraphrasing of an implicit aspect's label.}
\
\end{table}